\def\year{2021}\relax
\documentclass[letterpaper]{article} 
\usepackage{aaai21}  
\usepackage{times}  
\usepackage{helvet} 
\usepackage{courier}  
\usepackage[hyphens]{url}  
\usepackage{graphicx} 
\usepackage{natbib}  
\usepackage{caption} 
\usepackage{changepage}
\usepackage{setspace}
\usepackage{amsthm,amsmath,amssymb}
\usepackage{mathrsfs}
\usepackage{array}
\usepackage{colortbl}
\usepackage{arydshln}
\usepackage[ruled]{algorithm2e}
\usepackage{multirow}
\usepackage{colortbl}
\usepackage{bbding}
\usepackage[switch]{lineno}
\usepackage{footmisc} 
\definecolor{mygray}{gray}{.9}
\makeatletter
\newcommand{\thickhline}{%
    \noalign {\ifnum 0=`}\fi \hrule height 1pt
    \futurelet \reserved@a \@xhline
}

\title{Read, Retrospect, Select: An MRC Framework to Short Text Entity Linking}

\author{
    Yingjie Gu, \textsuperscript{\rm 1 \footnote{This work was done during an internship at Huawei.}}
    Xiaoye Qu, \textsuperscript{\rm 2}
    Zhefeng Wang, \textsuperscript{\rm 2 \textsuperscript{$\dagger$}}
    Baoxing Huai, \textsuperscript{\rm 2}
    Nicholas Jing Yuan, \textsuperscript{\rm 2}
    Xiaolin Gui \textsuperscript{\rm 1 \footnote{Corresponding authors.}}\\
}

\affiliations{
    \textsuperscript{\rm 1} Faculty of Electronics and Information Engineering, Xi’an Jiaotong University, Xi'an 710049, China \\
    \textsuperscript{\rm 2} Huawei Cloud \& AI, China \\
    gyj123wxc@stu.xjtu.edu.cn, xlgui@mail.xjtu.edu.cn, \\
    \{quxiaoye, wangzhefeng, huaibaoxing, nicholas.yuan\}@huawei.com
}

\begin{document}

\maketitle

\begin{abstract}

Entity linking (EL) for the rapidly growing short text (e.g. search queries and news titles) is critical to industrial applications. Most existing approaches relying on adequate context for long text EL are not effective for the concise and sparse short text. In this paper, we propose a novel framework called \textbf{M}ulti-turn \textbf{M}ultiple-choice \textbf{M}achine reading comprehension (\textbf{M3}) to solve the short text EL from a new perspective: a query is generated for each ambiguous mention exploiting its surrounding context, and an option selection module is employed to identify the golden entity from candidates using the query. In this way, M3 framework sufficiently interacts limited context with candidate entities during the encoding process, as well as implicitly considers the dissimilarities inside the candidate bunch in the selection stage. In addition, we design a two-stage verifier incorporated into M3 to address the commonly existed unlinkable problem in short text.
To further consider the topical coherence and interdependence among referred entities, M3 leverages a multi-turn fashion to deal with mentions in a sequence manner by retrospecting historical cues.
Evaluation shows that our M3 framework achieves the state-of-the-art performance on five Chinese and English datasets for the real-world short text EL. 

\end{abstract}

\section{Introduction}
The task of entity linking (EL), also known as entity disambiguation, aims to link a given mention that appears in a piece of text to the correct entity in a specific knowledge base (KB) \cite{shen2014entity}. 
Recently, with the explosive growth of short text in the web, entity linking for short text plays an important role in a wide range of industrial applications, such as parsing queries in search engines, understanding news titles and comments in social media (e.g. Twitter and Wechat). 

In this paper, we focus on short texts entity linking problem. Under this circumstance, the input text is composed of few tens of terms and the surrounding context of mentions is naturally scarce and concise \cite{ferragina2010tagme}. Traditional entity linking methods \cite{gupta2017entity, newman2018jointly} mainly employ encoded context and pre-trained candidate entity embeddings to assess topic level context compatibility for entity disambiguation. In this way, the disambiguation process degrades into a semantic matching and entity ranking problem. 
Specifically, recent state-of-the-art entity linking models \cite{gillick2019learning, logeswaran2019zero} score each pair of mention context and candidate entity based on their abstract representation. This means that the model lacks fine-grained interaction between mention context and their candidate entities. Semantic matching operations between them are performed only at the encoder output layer and are relatively superficial. Therefore, it is difficult for these models to capture all the lexical, semantic, and syntactic relations for precise entity disambiguation. 
Although these approaches that benefited from sufficient context achieve significant progress in long text, they fail to process sparse short text due to the inferiority of a restricted context model.  
Thus, it would be favorable if an entity linking framework can make full use of limited context. 


\begin{figure}[t]
\centering
\includegraphics[width=1.0\columnwidth]{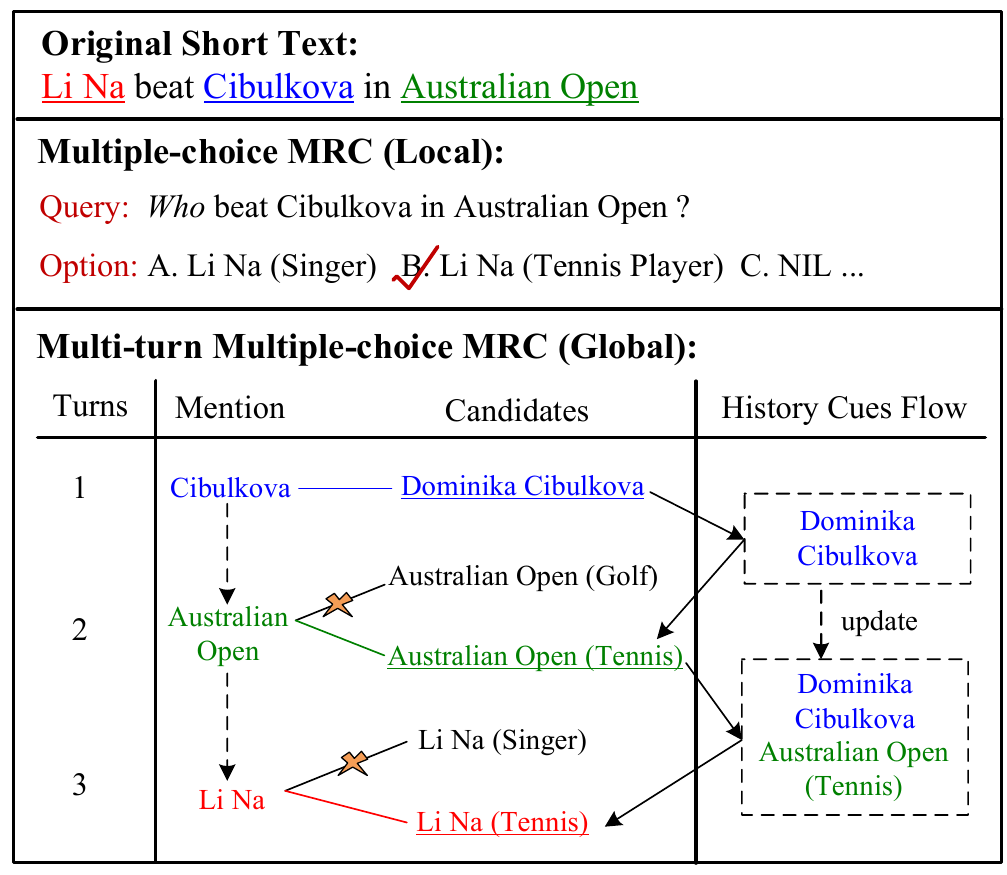} 
\caption{An illustration of our M3 framework for short text entity linking. The global disambiguation order depends on the ambiguity degrees.}
\label{fig1}
\end{figure}

To alleviate above issue, we propose a \textbf{M}ulti-turn \textbf{M}ulti-choice \textbf{M}achine reading comprehension (\textbf{M3}) framework to tackle the short text EL problem. 
As shown in Fig. \ref{fig1}, to identify the ambiguous mention ``\textit{Li Na}'' in the short text ``\textit{\underline{Li Na}} beat \textit{\underline{Cibulkova}} in \textit{\underline{Australian Open}}'', a query is generated using its surrounding context as below: 

\begin{spacing}{1.2}
\begin{adjustwidth}{0.3cm}{0.3cm}
\textit{Who beat Cibulkova} \textit{in Australian Open ?}
\end{adjustwidth}
\end{spacing}
Then, an option selection module is further employed to select the correct entity within the candidate bunch based on the given query. This formulation comes with the following key advantages: (1) Considering that the mention’s immediate context is a proxy of its type \cite{chen2020improving}, this form of query construction injects latent mention type information into the query embeddings. From the context ``\textit{beat Cibulkova}'', it is feasible for the model to infer the potential entity type of mention is person. 
(2) In the option selection stage, query and candidates obtain token-by-token interaction, achieving lexical and syntax similarities compared to concatenating mention and candidate embedding during MRC encoding.
(3) With the multi-choice setting, candidates are considered simultaneously, thus the comparison among candidates is comprehensive and the difference of scores among candidates is absolute, while the scores of different candidates are relative in both binary classification and ranking strategy. In Figure 1, the dissimilarities among candidates (``\textit{Singer}'' vs ``\textit{Tennis Player}'') implicitly attracts more attention and the shared part (``\textit{Li Na}'') between options is less highlighted compared to a binary classification strategy \cite{chengentity}.

In addition, short texts in web corpora are naturally time-sensitive and it is possible that some mention does not have its corresponding entity in the given KB. 
In order to solve these unlinkable mentions which are labeled as NIL (a special token \cite{shen2014entity}), we further design an NIL verifier incorporated into the M3 framework. Specifically, we devise a two-stage verification strategy: 1) The first stage yields a preliminary decision by sketchily reading the query. 2) The second stage verifies the concrete option and returns the final prediction via intensive reading. 
To further capture the relations among mentions for global disambiguation, we propose a multi-turn fashion to flexibly retrospect historical disambiguation cues in our M3 framework.
As shown in Fig. 1, when processing the mention ``\textit{Li Na}'', the candidate entity ``\textit{Li Na (Tennis Player)}'' shows strong relationship with last step referent entity ``\textit{Australian Open (Tennis)}''. This sample conforms to our motivation 
of M3 that knowledge from previously linked entities can be accumulated as dynamic context to facilitate later decisions. To alleviate the error propagation along the knowledge pass, we devise a controllable gate mechanism to prefer the most relevant entities. 
We conduct extensive experiments on both Chinese (Wechat and two out-domain test sets, CNDL Ex and Tencent News) and English (Webscope and an out-domain test KORE50) short text EL datasets. Our M3 framework achieves remarkable improvements of 5.62 percentage points on Wechat test set and an average of 5.03 percentage points on Webscope test set over five different runs compared with the state-of-the-art short text EL model \cite{chengentity}. In addition, we conduct detailed experimental analysis on Wechat dataset to show the effectiveness of our M3 framework.

Our contributions can be summarized as following: 
\begin{itemize}
\item To the best of our knowledge, it is the first attempt that a novel multi-turn multiple-choice machine reading comprehension (M3) framework is proposed for short text EL.
\item Our M3 framework devises an NIL verifier to handle unlinkable mention prediction in local disambiguation. Moreover, we design a multi-turn mechanism with a history flow for M3 to address the global disambiguation in short text. 
\item Experiments on five different datasets show that our M3 framework achieves state-of-the-art performance on both Chinese and English short text entity linking tasks.
\end{itemize}

\section{Related Work}

\begin{figure*}[t]
\centering
\includegraphics[width=2.0\columnwidth]{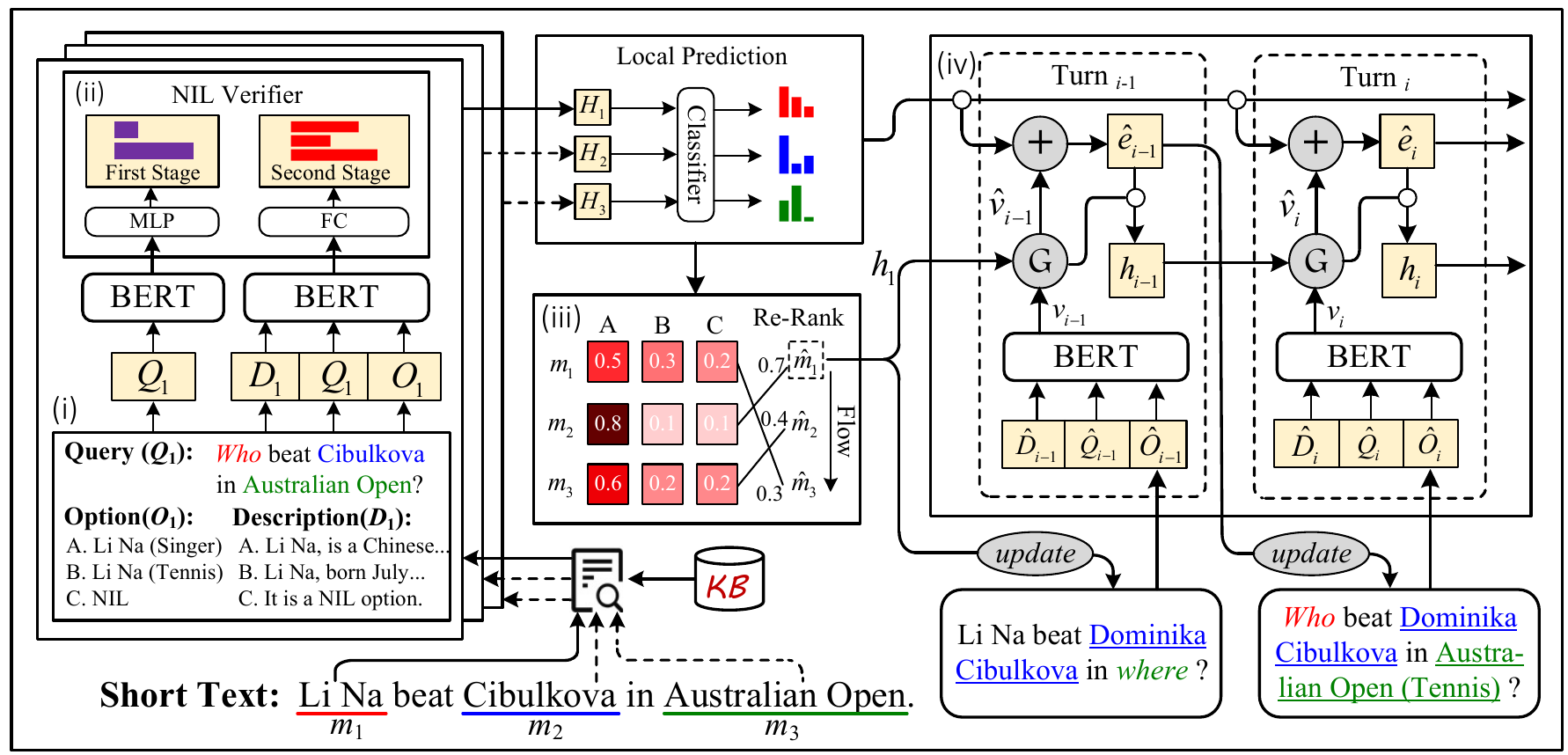}
\caption{Illustration of our proposed M3 framework for short text entity linking. (i) The input short text first forms multi-choice MRC with options from candidate entities and corresponding descriptions from KB for each mention. (ii) Then a NIL verifier to initially determine whether the mention is linkable jointly works with multi-choice MRC to conduct local prediction. (iii) Subsequently, all mentions are re-ranked for global disambiguation inputting sequence based on the local prediction confidence, i.e. the global decision order is $\hat{m}_1(m_2)\to \hat{m}_2(m_3)\to \hat{m}_3(m_1)$. (iv) Finally, a multi-turn module is devised to performed global disambiguation with updated query and entity information flow.}
\label{fig2}
\end{figure*}

\subsection{Short Text Entity Linking}
The task of entity linking (EL) is commonly formalized as an entity ranking task according to local and global similarity scores \cite{gupta2017entity, eshel2017named, chen2018short, gillick2019learning}. Local score of candidates is computed by the representations interacted by context of mention and candidate entity. Global EL deals with simultaneous disambiguation for all mentions in the whole text and produces global scores. Finally, candidate entities are ranked via integrated local and global scores and the candidate with highest score will be selected as the entity. 

There are a few previous works focused on short text entity linking. \citet{ferragina2010tagme} designed a system for short text to solve ambiguity and polysemy in the anchor-page mappings of Wikipedia. \citet{blanco2015fast} gave a solution to represent the entity as the centroid of word vectors of its relevant words. However, it makes the model hard to predict the true entity since relevant words of entities are noisy and representations by word vector are implicit. \citet{yang2015s} introduced a learning framework for short text EL, which combines non-linearity and efficiency of tree-based models with structured prediction. Afterwards, \citet{chen2018short} proposed a method that regards concepts of entities as explicitly fine-grained topics to solve the sparsity and noisy problem of short text entity linking. More recently, \citet{sakor2019old} proposed a tool to jointly solve the challenges of both entity and relation linking, while it is specifically designed for English dataset and not suitable for Chinese. \citet{chengentity} treated short text entity linking problem as a binary classification task with BERT encoder \cite{devlin2018bert}, which shows significant results and takes the first place in CCKS 2019 challenge. 
\subsection{Machine Reading Comprehension (MRC)}
Machine reading comprehenison (MRC) can be roughly categorized by cloze \cite{hill2015goldilocks}, multiple-choice \cite{lai2017race, clark2018think}, span-extraction \cite{ rajpurkar2018know} and generation \cite{nguyen2016ms} according to the answer types. 
Over the last few years, many tasks in natural language processing have been framed as reading comprehension while abstracting away the task-specific modeling constraints. For example, \citet{he2015question} showed that semantic role labeling tasks could be solicited by using question-answer pairs to represent the predicate-argument structure. In recent work, \citet{li2019unified} reformalized the NER task as an MRC question answering task, which is efficiently capable of addressing overlapping entity recognition problems through constructing the query contained prior knowledge. \citet{wu2019coreference} treated the co-reference resolution problem as an MRC task. The passage is split into sentences where marked with the proposal mentions and the objective is to find the answer set (i.e. co-reference clusters) from the whole passage according to the current question (i.e. sentence). 

\section{Method}
Our M3 framework for short text entity disambiguation mainly consists of two modules: Local Model with Multiple-choice MRC and NIL Verifier, and Global Model with Multi-turn MRC. The structure of M3 is shown in Fig. 2.

\subsection{Task Formalization and Candidates Generation}
Formally, given a short text $\mathcal{S}$ containing a set of identified mentions $M=\{m_1, m_2,...,m_n\}$. The goal of an entity linking system is to find a mapping that links each mention $m_i$ to a target entity $e_i$ which is an unambiguous page in a referent Knowledge Base (e.g. Baidu Baike for Chinese and Wikidata for English) or predict that there is no corresponding entity to current mention in the KB (i.e. $e_i$=NIL). 

Before entity disambiguation, for each mention $m_i$, potential candidate entities 
$\mathcal{O}_i\in\{e^1_i,...,e^{K}_i\}$ are first chosen by candidate generation from a specific KB, where $K$ is a pre-defined parameter to prune the candidate set. It is worth noting that each candidate $\mathcal{O}^j_i \in \mathcal{O}_i$ possesses one corresponding description $\mathcal{D}^j_i$ (as shown in Figure 2) in KB which serves as supporting descriptions.
Following previous works \cite{fang2019joint,le2019distant}, we adopt the surface matching method to generate candidate entities for each mention.  

\subsection{Local Model}
We formalize the local entity disambiguation as a multiple-choice MRC paradigm which consists of query construction and option selection module. Moreover, a NIL verifier is designed to facilitate NIL decision. 
In detail, to construct the query, we leverage pre-trained model BERT as our backbone. It is worth noting that the mention $m_i$ is replaced with a single [MASK] token instead of an explicit question word to construct the query $\mathcal{Q}_i$ in the actual scene.
For example, when identifying the mention ``\textit{Cibulkova}'' in  ``\textit{Li Na beat \underline{Cibulkova} in Australian Open}''. The query is described as below: 
\begin{spacing}{1.2}
\begin{adjustwidth}{0.3cm}{0.3cm}
\textit{Li Na beat} [MASK] \textit{in Australian Open.}
\end{adjustwidth}
\end{spacing}
\noindent  
As the explicit type information is not always available in the KB (for example Baidu-Baike in our paper), latent entity type (i.e. \textit{who}) can be perceived by reading the context around the [MASK] token \cite{chen2020improving}. Here we do not further incorporate the mention name into the query as it makes negligible performance difference.
Then, in the option selection process, for each option $\mathcal{O}^j_i$ of mention $m_i$, we further concatenate $\mathcal{D}^j_i$, $\mathcal{Q}_i$, and $\mathcal{O}^j_i$ with [CLS] and [SEP] tokens as the input sequence:
\begin{equation}
\mathcal{S}^j_i = \{{\rm [CLS]}\ \mathcal{D}^j_i\ {\rm [SEP]}\ \mathcal{Q}_i\ {\rm [SEP]}\ \mathcal{O}^j_i\ {\rm [SEP]}\}
\end{equation}
The target of option selection module is to select the correct answer (i.e. ground-truth entity) from the available options by making full use of their supporting knowledge.
For a query with $K$ answer options, we first obtain $K$ input queries: $\mathcal{S}^1_i,\mathcal{S}^2_i,...,\mathcal{S}^K_i$. Afterwards, we feed each query into BERT encoder and the final prediction for each option is obtained by a feed-forward layer with softmax function over the uppermost layer representation of BERT.
\begin{equation}
H_i^j={\rm BERT}(\mathcal{S}_i^j), j \in \{1,2,...,K\}
\end{equation}
\begin{equation}
\tilde{y}_i = {\rm Softmax}(W_1^TH_i + b_1)
\end{equation}
where $H_i\in\mathbb{R}^{d\times K}$, 
$W_1\in\mathbb{R}^d$ and $b_1\in\mathbb{R}^K$ are the learnable parameter and bias respectively. The training loss function of answer entity prediction is defined as cross entropy.
\begin{equation}
    \mathcal{L}_{ans} = -\frac{1}{N}\sum^N_{i=1}\sum^K_{j=1}y^j_i{\rm log}(\tilde{y}^j_i)
\end{equation}
where $\tilde{y}^j_i$ denotes the prediction of each answer and $y^j_i$ is the gold entity of $m_i$. $N$ is the number of examples.

\noindent \textbf{NIL Verifier}  
Considering the frequently appeared unlinkable problem in short text, following a natural practice of how humans solve linkable mention: the first step is to read through the query and obtain an initial judgement; then, people re-read the query and verify the answer if not so sure, we propose a two-stage verification mechanism. In the first stage, the preliminary judgment is determined by sketchy reading the query $\mathcal{Q}_i$. In specific, 
\begin{equation}
\bar{y}_i = \sigma({\rm MLP}({\rm BERT}(\mathcal{Q}_i)))
\end{equation}
here $\sigma$ is the sigmoid function, MLP denotes a multi-layer perception.
Then we adopt binary cross entropy loss to train this classifier. 
\begin{equation}
    \mathcal{L}_{nil} = -\frac{1}{N}\sum^N_{i=1}[y_i{\rm log}(\bar{y}_i)+(1-y_i){\rm log}(1-\bar{y}_i)]
\end{equation}
where $\bar{y}_i$ denotes the prediction and $y_i$ is the target indicating whether mention $m_i$ is linkable or not. $N$ is the number of training examples. 

Then, in the second stage, an additional option marked as ``\textit{NIL}'' (the corresponding description is ``This is a NIL option'') is appended to candidate entity set in the above local multi-choice model, which is capable to participate in the comparison between options. We argue the first stage is crucial because if only the second stage exists, the model will incline to the linkable options which share more components with the query. Finally, the joint loss for our local model incorporated with NIL verifier is the weighted sum of the answer loss and NIL loss.
\begin{equation}
    \mathcal{L}_{local} = \alpha_1\mathcal{L}_{ans} + \alpha_2\mathcal{L}_{nil}
\end{equation}
where $\alpha_1$ and $\alpha_2$ are two hyper-parameters that balance the weight of two losses. 

After local disambiguation, we obtain all candidate entity representations for each mention in the short text, which will be used for following global disambiguation. Meanwhile, the local scores calculated by Eq. 3 will be utilized for sequence ranking in the next stage.

\subsection{Global Model}

In this section, we aim to capture topical coherence and interdependency among all mentions for global disambiguation. To achieve this goal, we treat the global disambiguation as a multi-turn fashion extending from above multiple-choice paradigm. The intuitive idea is to utilize previously linked entities to enhance the later decisions through a dynamic multi-turn way. 
Before global disambiguation, according to the study \citet{yamada2020global}, starting with mentions that are easier to disambiguate will be effective to reduce the interference of noise data. In our full M3 framework, we rank mentions via their ambiguity degrees produced by the local model with below rules:
\begin{equation}
    \{\hat{m}_1,\hat{m}_2,...,\hat{m}_n\}=\text{Rank}\{\mathop{{\rm max}}\limits_{j,k \in K}{\lVert \tilde{y}^j_i-\tilde{y}^k_i\rVert}_{L_1}\}
\end{equation}

After sorting mentions, the representation of entity with highest score with respect to mention $\hat{m}_1$ is set to $h_1$, forming the initial history cue. Then, to obtain the global entity scores for following mention $\hat{m}_i$, we first utilize the linked entities to update current query $\hat{\mathcal{Q}}_{i}$, 
which is capable of accumulating previous knowledge in context level. For example, the query for identifying ``\textit{Australian Open}'' will be updated with the previously linked entities as following (``\textit{Cibulkova}'' has been linked to ``\textit{Dominika Cibulkova}''):
\begin{spacing}{1.2}
\begin{adjustwidth}{0.3cm}{0.3cm}
\textit{Li Na beat \underline{Dominika Cibulkova} in} [MASK].
\end{adjustwidth}
\end{spacing}
\noindent Then the updated query and each candidate entity of current mention are concatenated with special tokens [CLS] and [SEP] as the input sequence. Similar to the local model, we leverage the BERT encoder to obtain the representations of each candidate for current mention:
\begin{equation}
v^j_i = {\rm BERT}\{{\rm [CLS]}\ \hat{\mathcal{D}}^j_i\ {\rm [SEP]}\ \hat{\mathcal{Q}}_i\ {\rm [SEP]}\ \hat{\mathcal{O}}^j_i\ {\rm [SEP]}\}
\end{equation}
Subsequently, we propose to introduce historical cues for current mention disambiguation. However, some previously linked entities may be irrelevant to the current mention and several falsely linked entities may even lead to noise. For this purpose, a gate mechanism is desired to control which part of history cues should be inherited. 
In specific, a gated network is designed on the current and historical representations $h_{i-1}$ as follows:
\begin{equation}
u^j_i= \sigma(W_u[v^j_i;h_{i-1}])
\end{equation}
\begin{equation}
f^j_i={\rm tanh}(W_f[u^j_i \odot h_{i-1};v^j_i])
\end{equation}
\begin{equation}
g^j_i=\sigma(W_iv^j_i + W_hh_{i-1})
\end{equation}
\begin{equation}
\hat{v}^j_i=g^j_i \odot f^j_i + (1-g^j_i) \odot h_{i-1}
\end{equation}
where $W_u, W_f\in\mathbb{R}^{d\times 2d}, W_i, W_h\in\mathbb{R}^{d\times d}$ are learnable parameters.  
$f^j_i$ denotes the fusion of history information and current candidate input, and $g^j_i$ is the control gate to determine how much history information will be remained for each candidate entity. Finally, $\hat{v}^j_i$ will be output to calculate the global scores of candidates for the current mention.
\begin{equation}
\hat{y}^j_i={\rm Softmax}(W_2^T\hat{v}^j_i+b_2)
\end{equation}
\noindent where $W_2 \in\mathbb{R}^d$ and $b_2\in\mathbb{R}$ are the learnable parameter and bias respectively. After prediction, the representation of selected entity $\hat{e}_i$ with highest global score $\hat{y}^j_i$ for current mention $\hat{m}_i$ is flowed to next step of global disambiguation as updated historical representation. 
The training loss function for global model is defined as cross entropy.
\begin{equation}
    \mathcal{L}_{global} = -\frac{1}{N}\sum^N_{i=1}\sum^K_{j=1}y^j_i{\rm log}(\hat{y}^j_i)
\end{equation}
\noindent where $\hat{y}^j_i$ denotes the global prediction of answer and $y^j_i$ is the ground-truth entity of $m_i$. $N$ is the number of examples.

 \noindent \textbf{Rear Fusion} Rear fusion is the combination of predicted scores of local model and global model, determining the final disambiguated entity list for each mention in short text.
 \begin{equation}
    \text{Score}(e^j_i) = \beta\tilde{y}^j_i + (1 - \beta)\hat{y}^j_i
\end{equation}
where $\beta$ is the weight to balance local and global score. \\ 
The details of our model are presented in Algorithm 1.

\begin{algorithm}[t]\footnotesize
    \caption{M3 framework for Short Text EL}
    \KwIn{Short text $\mathcal{S}$ with $M=\{m_1,...,m_n\}$, candidates $\mathcal{O}_m$ and descriptions $\mathcal{D}_m$ for each mention $m$}
    \KwOut{Linked entities $E=\{e_1,...,e_n\}$ for all mentions}
    \For{m {\rm in} $M$}
    {
    Build Multi-choice MRC paradigm $[\mathcal{D}_m,\mathcal{Q}_m,\mathcal{O}_m]$  \;
    Compute $\mathcal{L}_{ans}$ for all options with Eq.($1\sim 4$)\;
    Compute $\mathcal{L}_{nil}$ by NIL Verifier with Eq.($5\sim 6$)\;
    $ \mathcal{L}_{local} = \alpha_1\mathcal{L}_{ans} + \alpha_2\mathcal{L}_{nil}\ $ \;
    Update parameters with joint loss $ \mathcal{L}_{local}$ \;
    }
    Re-rank mention $\hat{M}=\{\hat{m}_1,...,\hat{m}_n\}$ with Eq.($8$)\;
    \For{$i \; {\rm in} \; 1\ {\rm to}\ n$}{
	\eIf{$i==1$}
	{
	{\rm Select $\hat{e}_1\ {\rm in}\ \hat{m}_1$ by local score}\; 
	{Obtain initial history vector $v_{\hat{e}_1}\ {\rm as}\ h_1$}\;
	}
	{
	Update Query by linked entity $\hat{\mathcal{Q}}_{i-1}\stackrel{\hat{e}_{i-1}}{\longrightarrow}\hat{\mathcal{Q}}_{i}$  \;
	Build Multi-turn MRC paradigm $[\mathcal{D}_{\hat{m}_i},\hat{\mathcal{Q}}_{i},\mathcal{O}_{\hat{m}_i}]$\;
	Obtain $v_i$ for all options with Eq.($9$)\;
	Input $v_i,h_{i-1}$ to Gated Network, and select the target entity $\hat{e}_i$ for $\hat{m}_i$ by Eq.($10\sim 14$)\;
	Update the history vector $v_{\hat{e}_i}\to h_i$\;
	Update parameters with loss $\mathcal{L}_{global}$ by Eq.($15$)\;
	}
}
\end{algorithm}

\section{Experiments}
\subsection{Datasets}
In order to verify the effectiveness of our framework, we conduct experiments on five Chinese and English datasets considering both in-domain and out-domain settings. Since most of the existing datasets on EL are based on long text, which are not suitable for the task of short text EL. We find two public English datasets: Webscope \cite{blanco2015fast} and KORE50 \cite{hoffart2012kore} that are suitable for short text EL. However, there are few existing datasets for Chinese short text EL with high quality annotation. Therefore, we construct two Chinese datasets (Wechat and Tencent News) for short text EL and expand one public dataset CNDL \cite{chen2018short} to CNDL Ex. Due to the limited space, the detailed construction process and short text samples will be available in the supplementary material. The statistics of these datasets are shown in Table 1. To make a fair comparison, we train all models on Webscope for English and test on Webscope and KORE50. For Chinese datasets, we train on Wechat dataset and test them on the Wechat, CNDL Ex, and Tencent News dataset.

\begin{table}[t]\small
\centering
\begin{tabular}{lcccc}
\thickhline
Dataset  & Text & Language & Avg./men. & \% NIL \\ \hline
Wechat          & 11,439  & Cn          & 2.00 & 14.34\%    \\
CNDL Ex           & 877    & Cn            & 2.03 & 16.50\%    \\
Tencent News       & 1,000   & Cn            & 1.73 & 9.71\%   \\ 
Webscope           & 2,635   & En            & 2.26 & -  \\
KORE50             & 50     & En            & 2.88 & -   \\\thickhline
\end{tabular}
\caption{Statistics of short text EL datasets. \% NIL is the percentage of NIL mentions in all mentions.}
\end{table}

\begin{table*}[t]\small
\centering
\scalebox{0.93}{
\begin{tabular}{lccccccc}
\thickhline
Model  & Wechat  & CNDL Ex* & Tencent News*  & Avg Cn & Webscope & KORE50* & Avg En.\\ \hline
$\hat{p}(e|m)$ \cite{chen2018short}                & $60.68$   & $61.24$ & $60.92$ & $60.95$ & $51.71$ & $50.00$ & $50.86$\\ 
\cdashline{1-8}[1pt/2pt]
\citet{eshel2017named}             & $77.68$ & $75.59$ & $79.94$ & $77.74$ & $84.58$ & $46.03$ & $65.31$\\
\citet{kolitsas2018end}             & $78.18$ & $74.97$ & $79.48$ & $77.54$ & $87.08$ & $53.17$ & $70.13$\\
\citet{shahbazi2019entity}              & $78.26$ & $75.37$ & $79.19$ & $77.61$ & $82.22$ & $57.94$ & $70.08$\\
\citet{chen2020improving}              & $81.17$ & $78.09$ & $82.77$ & $80.68$& $84.17$ & $63.49$ & $73.83$ \\
\citet{chengentity}  $^{\clubsuit}$         & $89.20$   & $89.15$ & $88.32$  & $88.89$ & $87.45$ & $71.58$ & $79.52$ \\ 
\hline
M3 (base) $^{\clubsuit}$               & $93.07\pm0.4$   & $91.75\pm0.5$ & $89.27\pm0.5$ & $91.36$ & $89.24\pm0.5$ & $71.42\pm0.9$  & $80.33$\\
M3 (local) $^{\clubsuit}$         & $94.06\pm0.2$   & $93.37\pm0.4$ & $89.86\pm0.5$ & $92.43$ & $89.24\pm0.5$ & $71.42\pm0.9$  & $80.33$\\
M3 (full) $^{\clubsuit}$ & $\textbf{94.82}\pm0.1$   & $\textbf{94.33}\pm0.2$ & $\textbf{90.57}\pm0.4$& $\textbf{93.24}$ & $\textbf{92.48}\pm0.5$ & $\textbf{74.28}\pm0.5$& $\textbf{83.38}$ \\ 
\thickhline
\end{tabular}}
\caption{Evaluation on both Chinese and English datasets. ${\clubsuit}$ indicates methods specifically designed for short texts. * denotes datasets only for transfer evaluation. M3 (base) is a variant of M3 (local) model without NIL Verifier. M3 (full) consists of both local and global model. On the English datasets, M3 (base) is equal to M3 (local) as there are no NIL mentions within them.
}
\end{table*}

\begin{table}[t]\small
\centering
\scalebox{0.95}{
\begin{tabular}{l|l|cc}
\thickhline
\multirow{2}{*}{Components}   & \multirow{2}{*}{Module} & \multicolumn{2}{c}{Accuracy} \\ \cline{3-4} 
                              &                         & Total         & $\Delta$            \\ \hline \hline
All                     & M3 (full)               & 94.92         & -            \\
\hline
NIL Verifier                  & without NIL Verifier  & 93.76         & 1.16         \\ \hline
Decision Order                & without mention re-ranking           & 94.68         & 0.24         \\ \hline
Query update                        & without query update       & 94.43         & 0.49         \\\hline
\multirow{2}{*}{Gate Mechanism} & Replace with Concatenate        & 94.56         & 0.36         \\
                              & Replace with GRU                & 94.64         & 0.28         \\\hline
History Flow                  & Replace with last history       & 94.48         & 0.44         \\
\thickhline
\end{tabular}}
\caption{Ablation study of M3 framework on Wechat dataset.}
\end{table}


\subsection{Experiment Setup}
In this paper, our goal is to demonstrate the superiority of M3 framework for short text entity linking. As a full EL system in real-world includes mention detection, candidate generation, and entity disambiguation. Here we focus on disambiguation method and fairly comparing with previous methods under same preliminary condition, i.e. same candidate generation strategy. In detail, We adopt alias dictionary look-up to search the candidates from the Baidu-Baike (Baidu Baike on Feb. 2019) on Chinese datasets and rank them by their page-view in the knowledge base. For English dataset, we adopt surface matching methods to match Wikipedia page (Wikipedia Dump on May. 2020) and calculate the prior probability between mention and candidate description for ranking. Considering the NIL problem, we retain top K candidates, and then K candidates and NIL token serve as candidate sets for each mention.

In our experiment, we leverage the pre-trained uncased BERT-Base model with 768 dimensions hidden representation as our backbone. For local model, we adopt Adam as optimizer with warmup rate 0.1, initial learning rate 5e-6, and maximum sequence length 256. The hyper-parameter $\alpha_1$, $\alpha_2$ are set to 0.75, 0.25. For the global encoder, we use Adam optimizer with a learning rate of 1e-5 and maximum of 512 tokens after tokenization. For each mention, the candidates' number $K$ is set to 5. In rear fusion, $\beta$ is 0.5. All experiments are performed on one Tesla P100 with 16G GPU memory. 

\subsection{Comparison Methods}
In our experiments, we compare our proposed model with the following baseline methods: 
(1) \citet{chen2018short} provided a mention-entity prior statistical estimation from the entity description in knowledge base. 
(2) \citet{eshel2017named} proposed to use a modified GRU to encode left and right context of a mention, which is capable of handling the EL problem with noisy local context. 
(3) \citet{kolitsas2018end} used bidirectional LSTM networks on the top of learnable char embeddings and represent entity mention as a combination of LSTM hidden states included in the mention spans.
(4) \citet{shahbazi2019entity} first proposed a method to learn an entity-aware extension of pre-trained ELMo \cite{peters2018deep} and obtains significant improvements in many long text EL tasks.
(5) \citet{chen2020improving} proposed to improve EL performance via capturing latent entity type information with BERT. This model is able to correct most of the type errors and obtains the state-of-the-art performance on lots of long text EL datasets.
(6) \citet{chengentity} proposed a strong method to handle Chinese short text EL and achieve first place in CCKS 2019 short text EL track. 
They treat the short text EL as a binary classification task and leverage BERT as a backbone to obtain deeply interactive representations between the mention context and candidate entities.
\begin{table}[tp]\small
\scalebox{0.95}{
\begin{tabular}{p{1.4cm}p{6.0cm}}
\thickhline
Text 1    & Remake from the 2005 Japanese TV series \textbf{``Queen's classroom"}.    \\
\multirow{2}{*}{Candidates} & (1) Queen's classroom (Japanese TV) \\
 & (2) Queen's classroom (Korea TV)\\
Binary  &  (1) 0.725 \ \ \ \ (2) \textbf{0.728} \XSolidBrush      \\
M3 (base)      & (1) \textbf{0.82} \CheckmarkBold \ \ (2) 0.18                                                         \\ \hline
Text 2    & Three episodes of documentary "\textbf{Masters In Forbidden City}" was popular all over country.                                                          \\
\multirow{2}{*}{Candidates} & (1) Masters In Forbidden City (Documentary) \\
 & (2) Masters In Forbidden City (Movie)\\
Binary  & (1) 0.951 \ \ \ \ (2) \textbf{0.953} \XSolidBrush   \\
M3 (base)     & (1) \textbf{0.69} \CheckmarkBold \ \ (2) 0.30  \\ \thickhline                                              
\end{tabular}}
\caption{Examples that wrongly predicted by binary model \citet{chengentity} but correctly predicted by M3 (base).}
\end{table}

\subsection{Results}
We present the entity linking evaluation results in Table 2. From Table 2 we can observe that compared to the strong baseline approaches based on the same BERT-based encoder, our M3 framework exhibits the state-of-the-art performance on both Chinese and English short text datasets. On the Wechat datasets, our M3 (local) achieves 4.86 percentage absolute improvement in terms of accuracy over the strong method \cite{chengentity} specifically designed for short text. Equipped with global module, the average performance of our model M3 (full) further increases to 94.82, indicating that our design successfully tackles mention coherence problems. On the Webscope dataset, our M3 (full) model also obtains incredible improvement compared to baseline EL approaches \cite{chengentity, chen2020improving} by 5.03, 8.31 respectively. 
In addition, Table 2 shows the performance on two Chinese out-domain and one English out-domain datasets. On CNDL Ex and Tencent News datasets, our model presents consistent performance improvement by 5.18 and 2.25 percentage in terms of accuracy compared to SOTA method \cite{chengentity}. On KORE50 dataset, the baseline \cite{chengentity} performs slightly better than our M3 (local) model by 0.16 but far worse than our M3 (full) model by 2.7 percentage. The performance on all these out-domain datasets reveals the robustness of our model.

\subsection{Ablation Study}
\begin{table*}[ht]\small
\centering
\scalebox{0.93}{
\begin{tabular}{cllcccl}
\thickhline
\rowcolor{mygray}
\multicolumn{7}{l}{\begin{tabular}[c]{@{}l@{}}
\textbf{Short text (Wechat)}: According to the different ethnic characteristics of \textbf{\underline{Terrans}}, \textbf{\underline{Zerg}} and \textbf{\underline{Protoss}}, mining, replenishing soldier,\\ and making all-round strategic decisions are carried out; after all, they have no divine shield skill of \textbf{\underline{Paladins}} in \textbf{\underline{World of Warcraft}}.\end{tabular}} \\ \hline
Turn                              & Mention                                         & Candidates                                              & Local                & Global               & Final & Golden Entity \\ \hline
1                                 & World of Warcraft                                      & World of Warcraft                                        & 0.99                       & -           & - &   World of Warcraft             \\ \hline
\multirow{2}{*}{2}                & \multirow{2}{*}{Zerg}                & (1) Zerg (StarshipTroopers)                & 0.08                       & 0.08  & 0.08 &   \multirow{2}{*}{Zerg (StarCraft)}                       \\
                                  &                                                 & (2) Zerg (StarCraft)                  & 0.91                       & 0.89       & 0.90 &                    \\ \hline
\multirow{2}{*}{3}                & \multirow{2}{*}{Protoss}                        & (1) Protoss (StarCraft)                                        & 0.28                       & 0.59      & 0.44 &   \multirow{2}{*}{Protoss (StarCraft)}                   \\
                                  &                                                 & (2) Protoss (Slayers)                                   &  \textbf{0.63}                       & 0.08    & 0.36 &                  \\\hline
\multirow{2}{*}{4}                & \multirow{2}{*}{Paladin}                        & (1) Paladin (Dungeon \& Fighter)                                        & 0.07                       & 0.01         & 0.04 &   \multirow{2}{*}{Paladin (World of Warcraft)}                \\
                                  &                                                 & (2) Paladin (World of Warcraft)                                   & 0.54                       & 0.97         & 0.76 &             \\\hline
\multirow{3}{*}{5} & \multirow{3}{*}{Terrans} & (1) Terrans (Warcraft) & 0.30 & 0.17& 0.24 &   \multirow{3}{*}{Terrans (StarCraft)}  \\
                    &                     & (2) Terrans (StarCraft) & 0.21 & 0.47 & 0.34 &    \\
                    &                     & (3) Terrans (Biological category) & \textbf{0.33} & 0.20 & 0.27 &    \\ \thickhline
\end{tabular}}
\caption{An example of our global model with multi-turn fashion to handle topical coherence in a short text with five mentions. The bold item is mistakenly predicted by local model but corrected by global model.}
\end{table*}

\begin{table}[h]\small
\centering
\begin{tabular}{lp{5.3cm}}
\thickhline
Text 1  & The neighbor called the police for \textbf{\underline{Bonnie}}, but the man pulled out a fruit knife that had been prepared for a long time. \\
M3 (base)        & Yu Hanmi (Actress)                                                                                                      \\
M3 (local)    & NIL                                                                                                                    \\ \hline
Text 2 & \textbf{\underline{Jiang Shan}}, who knocked in, stood in front of the principal's desk, who was sitting there waiting for him.               \\
M3 (base)        & Jiang Shan City (Zhejiang Province)                                                                                      \\
M3 (local)   & NIL                                                                                                                     \\ \thickhline
\end{tabular}
\caption{Examples of NIL Verifier.}
\end{table}

To better evaluate the contribution of various components to the overall performance, we conduct abundant ablation studies of M3 framework on Wechat dataset in Table 3. From the results, we can observe that: (1) When removing the NIL verifier the performance drops by 1.16, which indicates that our NIL verifier can effectively tackle NIL problem. (2) When linking mentions by the natural order of appearance in the text instead of re-ranking mention with Eq. 8, the result becomes worse and it means re-ranking helps improve linking ability. (3) If we do not update the query at each turn with previously linked entities, the performance will drop by 0.49, revealing that prior linked knowledge plays a significant role in global model. (4) Replacing our devised gate mechanism with a simple concatenate operation or vanilla GRU structure both degrades the performance, which denotes that our gate mechanism is more efficient to filter noisy entity information. (5) When only utilizing last step entity information as the history cue, the accuracy drops by 0.44, which shows that our history flow among all mentions contributes more to topical coherence capturing. 

\subsection{Analysis} We demonstrate the effectiveness of our proposed model from the following aspects:
\begin{itemize}
\item Does the multiple-choice paradigm in M3 framework better interact limited context with entities than binary classification \cite{chengentity}?
\item Does NIL Verifier facilitate discriminating NIL entities?
\item Can global model correct errors occurred in the local model with resort to topical coherence among mentions?
\item Can multi-turn strategy boost short texts entity linking with different number of mentions?
\end{itemize}

\subsubsection{Effectiveness of Multiple-choice Paradigm}
In Table 2, we present the results of our M3 (base) and \cite{chengentity} which treat the EL as a binary classification. From the result, it is obvious that our multi-choice strategy performs significantly better than binary classification in this task. In addition, 
we show some typical cases in Table 4 where the two candidate entities with similar descriptions are highly ambiguous. In this scenario, \cite{chengentity} which independently assigns scores to each candidate can barely distinguish the entities. Nevertheless, our M3 (base) successfully captures the micro-difference between candidates in the lexical-level and recognizes the true entity,  
which denotes that the dissimilarities among candidates attract more attention with our multiple-choice setting.

\subsubsection{Effectiveness of NIL Verifier}
As shown in Table 2 (M3 (base) vs M3 (local)) and Table 3 (NIL Verifier), NIL Verifier presents significant improvements in both scenarios. Moreover, we provide qualitative analyses to highlight the importance of NIL Verifier in Table 6. As shown in this table, when removing NIL verifier, the model tends to link the mention to intrinsic entities in the KB. 

\subsubsection{Topical Coherence Correction}
As shown in Table 2, M3 (full) achieves better results than M3 (local) with global model, especially on English datasets. To have an intuitive observation of the concrete process of global model, we also provide a prediction example on Wechat dataset in Table 5. According to this table, the multi-turn strategy effectively corrects several coherence errors with the help of historical information of linked entities. For example, when linking ``Protoss'' and ``Terrans'', 
the collective inference with linked history cue ``Zerg (StarCraft)'' promotes our global model to select an entity with highest topical coherence. 

\subsubsection{Generalization of Multi-turn Strategy}
Fig. 3 demonstrates the performance comparison on the short texts with different number of mentions. In total, our M3 (full) consistently performs better compared with model M3 (local). For the in-domain dataset (Figure 3 Left), our M3 (full) equipped with multi-turn module achieves an improvement of average 1.65\%. Especially on the text with 5 mentions, our global model gain the highest 2.1\% improvement. For the out-domain dataset (Figure 3 Right), our multi-turn structure obtains about 1.68\% average accuracy improvement by using the history cues. Different from in-domain dataset, the best improvement 2.6\% has been achieved by our multi-turn model on the text with 2 mentions. From both in-domain and out-domain settings, our global model with multi-turn strategy shows significant generalization.

\begin{figure}[t]
\centering
\includegraphics[width=1.0\columnwidth]{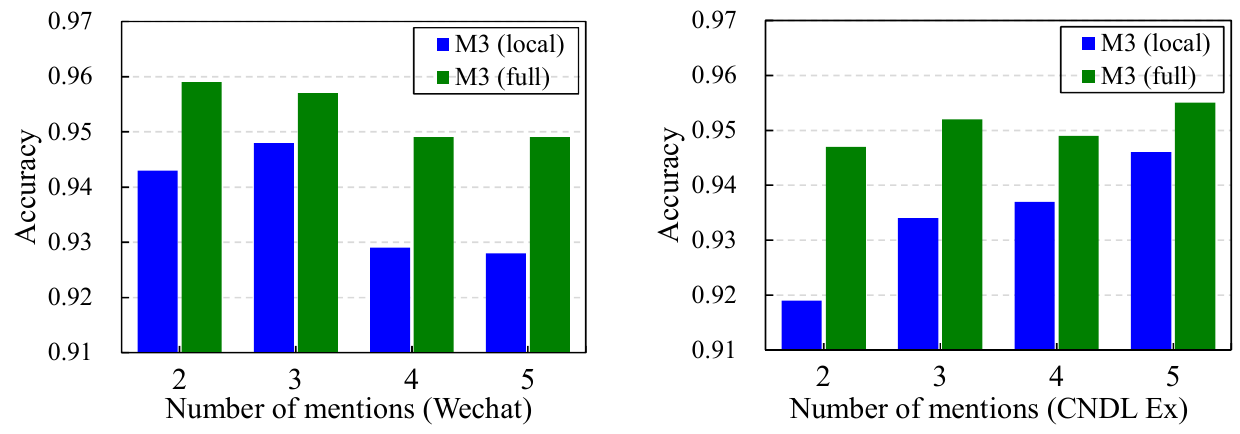} 
\caption{Results of global model for different number of mentions. Left: Wechat dataset. Right: CNDL EX dataset.}
\label{fig3}
\end{figure}

\section{Conclusion}
In this article, we presents a novel \textbf{M}ulti-turn \textbf{M}ultiple-choice \textbf{M}RC (M3) framework for short text EL. Firstly, we build query construction and option selection module for local disambiguation with an auxiliary NIL verifier for handling unlinkable entity problem. Then we leverage a multi-turn way with historical cues flow to tackle global topical coherence problem among mentions. The experiment results have proved that our M3 framework achieves the state-of-the-art performance on five Chinese and English short text datasets for real-world applications.
In fact, our M3 framework can integrate more types of information if it is available in the KB, such as relations between entities or explicit entity type information. Due to the restriction of the KB, we leave it as a future work.   

\section{Acknowledgments}
This paper was partially supported by National Key Research and Development Program of China (2018YFB1800304); Key Development Program in Shaanxi Province of China (2019GY-005, 2017ZDXM-GY-011)

\bibstyle{aaai21}
\bibliography{ref}

\begin{thebibliography}{29}
\providecommand{\natexlab}[1]{#1}
\providecommand{\url}[1]{\texttt{#1}}
\providecommand{\urlprefix}{URL }
\expandafter\ifx\csname urlstyle\endcsname\relax
  \providecommand{\doi}[1]{doi:\discretionary{}{}{}#1}\else
  \providecommand{\doi}{doi:\discretionary{}{}{}\begingroup
  \urlstyle{rm}\Url}\fi

\bibitem[{Blanco, Ottaviano, and Meij(2015)}]{blanco2015fast}
Blanco, R.; Ottaviano, G.; and Meij, E. 2015.
\newblock Fast and space-efficient entity linking for queries.
\newblock In \emph{Proceedings of the Eighth ACM International Conference on
  Web Search and Data Mining}, 179--188.

\bibitem[{Chen et~al.(2018)Chen, Liang, Xie, and Xiao}]{chen2018short}
Chen, L.; Liang, J.; Xie, C.; and Xiao, Y. 2018.
\newblock Short text entity linking with fine-grained topics.
\newblock In \emph{Proceedings of the 27th ACM International Conference on
  Information and Knowledge Management}, 457--466.

\bibitem[{Chen et~al.(2020)Chen, Wang, Jiang, and Lin}]{chen2020improving}
Chen, S.; Wang, J.; Jiang, F.; and Lin, C.-Y. 2020.
\newblock Improving entity linking by modeling latent entity type information.
\newblock \emph{arXiv preprint arXiv:2001.01447} .

\bibitem[{Cheng et~al.(2019)Cheng, Pan, Dang, Yang, Guo, Zhang, and
  Zhang}]{chengentity}
Cheng, J.; Pan, C.; Dang, J.; Yang, Z.; Guo, X.; Zhang, L.; and Zhang, F. 2019.
\newblock Entity Linking for Chinese Short Texts Based on BERT and Entity Name
  Embeddings.
\newblock \emph{China Conference on Knowledge Graph and Semantic Computing} .

\bibitem[{Clark et~al.(2018)Clark, Cowhey, Etzioni, Khot, Sabharwal, Schoenick,
  and Tafjord}]{clark2018think}
Clark, P.; Cowhey, I.; Etzioni, O.; Khot, T.; Sabharwal, A.; Schoenick, C.; and
  Tafjord, O. 2018.
\newblock Think you have solved question answering? try arc, the ai2 reasoning
  challenge.
\newblock \emph{arXiv preprint arXiv:1803.05457} .

\bibitem[{Devlin et~al.(2018)Devlin, Chang, Lee, and
  Toutanova}]{devlin2018bert}
Devlin, J.; Chang, M.-W.; Lee, K.; and Toutanova, K. 2018.
\newblock Bert: Pre-training of deep bidirectional transformers for language
  understanding.
\newblock \emph{arXiv preprint arXiv:1810.04805} .

\bibitem[{Eshel et~al.(2017)Eshel, Cohen, Radinsky, Markovitch, Yamada, and
  Levy}]{eshel2017named}
Eshel, Y.; Cohen, N.; Radinsky, K.; Markovitch, S.; Yamada, I.; and Levy, O.
  2017.
\newblock Named Entity Disambiguation for Noisy Text.
\newblock In \emph{Proceedings of the 21st Conference on Computational Natural
  Language Learning (CoNLL 2017)}, 58--68.

\bibitem[{Fang et~al.(2019)Fang, Cao, Li, Zhang, Zhang, and
  Liu}]{fang2019joint}
Fang, Z.; Cao, Y.; Li, Q.; Zhang, D.; Zhang, Z.; and Liu, Y. 2019.
\newblock Joint entity linking with deep reinforcement learning.
\newblock In \emph{The World Wide Web Conference}, 438--447.

\bibitem[{Ferragina and Scaiella(2010)}]{ferragina2010tagme}
Ferragina, P.; and Scaiella, U. 2010.
\newblock Tagme: on-the-fly annotation of short text fragments (by wikipedia
  entities).
\newblock In \emph{Proceedings of the 19th ACM international conference on
  Information and knowledge management}, 1625--1628.

\bibitem[{Gillick et~al.(2019)Gillick, Kulkarni, Lansing, Presta, Baldridge,
  Ie, and Garcia-Olano}]{gillick2019learning}
Gillick, D.; Kulkarni, S.; Lansing, L.; Presta, A.; Baldridge, J.; Ie, E.; and
  Garcia-Olano, D. 2019.
\newblock Learning Dense Representations for Entity Retrieval.
\newblock In \emph{Proceedings of the 23rd Conference on Computational Natural
  Language Learning (CoNLL)}, 528--537.

\bibitem[{Gupta, Singh, and Roth(2017)}]{gupta2017entity}
Gupta, N.; Singh, S.; and Roth, D. 2017.
\newblock Entity Linking via Joint Encoding of Types, Descriptions, and
  Context.
\newblock In \emph{Proceedings of the 2017 Conference on Empirical Methods in
  Natural Language Processing}, 2681--2690.

\bibitem[{He, Lewis, and Zettlemoyer(2015)}]{he2015question}
He, L.; Lewis, M.; and Zettlemoyer, L. 2015.
\newblock Question-answer driven semantic role labeling: Using natural language
  to annotate natural language.
\newblock In \emph{Proceedings of the 2015 conference on empirical methods in
  natural language processing}, 643--653.

\bibitem[{Hill et~al.(2015)Hill, Bordes, Chopra, and
  Weston}]{hill2015goldilocks}
Hill, F.; Bordes, A.; Chopra, S.; and Weston, J. 2015.
\newblock The goldilocks principle: Reading children's books with explicit
  memory representations.
\newblock \emph{arXiv preprint arXiv:1511.02301} .

\bibitem[{Hoffart et~al.(2012)Hoffart, Seufert, Nguyen, Theobald, and
  Weikum}]{hoffart2012kore}
Hoffart, J.; Seufert, S.; Nguyen, D.~B.; Theobald, M.; and Weikum, G. 2012.
\newblock KORE: keyphrase overlap relatedness for entity disambiguation.
\newblock In \emph{Proceedings of the 21st ACM international conference on
  Information and knowledge management}, 545--554.

\bibitem[{Kolitsas, Ganea, and Hofmann(2018)}]{kolitsas2018end}
Kolitsas, N.; Ganea, O.-E.; and Hofmann, T. 2018.
\newblock End-to-End Neural Entity Linking.
\newblock In \emph{Proceedings of the 22nd Conference on Computational Natural
  Language Learning}, 519--529.

\bibitem[{Lai et~al.(2017)Lai, Xie, Liu, Yang, and Hovy}]{lai2017race}
Lai, G.; Xie, Q.; Liu, H.; Yang, Y.; and Hovy, E. 2017.
\newblock RACE: Large-scale ReAding Comprehension Dataset From Examinations.
\newblock In \emph{Proceedings of the 2017 Conference on Empirical Methods in
  Natural Language Processing}, 785--794.

\bibitem[{Le and Titov(2019)}]{le2019distant}
Le, P.; and Titov, I. 2019.
\newblock Distant Learning for Entity Linking with Automatic Noise Detection.
\newblock In \emph{Proceedings of the 57th Annual Meeting of the Association
  for Computational Linguistics}, 4081--4090.

\bibitem[{Li et~al.(2019)Li, Feng, Meng, Han, Wu, and Li}]{li2019unified}
Li, X.; Feng, J.; Meng, Y.; Han, Q.; Wu, F.; and Li, J. 2019.
\newblock A unified mrc framework for named entity recognition.
\newblock \emph{arXiv preprint arXiv:1910.11476} .

\bibitem[{Logeswaran et~al.(2019)Logeswaran, Chang, Lee, Toutanova, Devlin, and
  Lee}]{logeswaran2019zero}
Logeswaran, L.; Chang, M.-W.; Lee, K.; Toutanova, K.; Devlin, J.; and Lee, H.
  2019.
\newblock Zero-Shot Entity Linking by Reading Entity Descriptions.
\newblock In \emph{Proceedings of the 57th Annual Meeting of the Association
  for Computational Linguistics}, 3449--3460.

\bibitem[{Newman-Griffis, Lai, and Fosler-Lussier(2018)}]{newman2018jointly}
Newman-Griffis, D.; Lai, A.~M.; and Fosler-Lussier, E. 2018.
\newblock Jointly Embedding Entities and Text with Distant Supervision.
\newblock In \emph{Proceedings of The Third Workshop on Representation Learning
  for NLP}, 195--206.

\bibitem[{Nguyen et~al.(2016)Nguyen, Rosenberg, Song, Gao, Tiwary, Majumder,
  and Deng}]{nguyen2016ms}
Nguyen, T.; Rosenberg, M.; Song, X.; Gao, J.; Tiwary, S.; Majumder, R.; and
  Deng, L. 2016.
\newblock Ms marco: A human-generated machine reading comprehension dataset .

\bibitem[{Peters et~al.(2018)Peters, Neumann, Iyyer, Gardner, Clark, Lee, and
  Zettlemoyer}]{peters2018deep}
Peters, M.~E.; Neumann, M.; Iyyer, M.; Gardner, M.; Clark, C.; Lee, K.; and
  Zettlemoyer, L. 2018.
\newblock Deep contextualized word representations.
\newblock In \emph{Proceedings of NAACL-HLT}, 2227--2237.

\bibitem[{Rajpurkar, Jia, and Liang(2018)}]{rajpurkar2018know}
Rajpurkar, P.; Jia, R.; and Liang, P. 2018.
\newblock Know What You Don’t Know: Unanswerable Questions for SQuAD.
\newblock In \emph{Proceedings of the 56th Annual Meeting of the Association
  for Computational Linguistics (Volume 2: Short Papers)}, 784--789.

\bibitem[{Sakor et~al.(2019)Sakor, Mulang, Singh, Shekarpour, Vidal, Lehmann,
  and Auer}]{sakor2019old}
Sakor, A.; Mulang, I.~O.; Singh, K.; Shekarpour, S.; Vidal, M.~E.; Lehmann, J.;
  and Auer, S. 2019.
\newblock Old is gold: linguistic driven approach for entity and relation
  linking of short text.
\newblock In \emph{Proceedings of the 2019 Conference of the North American
  Chapter of the Association for Computational Linguistics: Human Language
  Technologies, Volume 1 (Long and Short Papers)}, 2336--2346.

\bibitem[{Shahbazi et~al.(2019)Shahbazi, Fern, Ghaeini, Obeidat, and
  Tadepalli}]{shahbazi2019entity}
Shahbazi, H.; Fern, X.~Z.; Ghaeini, R.; Obeidat, R.; and Tadepalli, P. 2019.
\newblock Entity-aware ELMo: Learning Contextual Entity Representation for
  Entity Disambiguation.
\newblock \emph{arXiv preprint arXiv:1908.05762} .

\bibitem[{Shen, Wang, and Han(2014)}]{shen2014entity}
Shen, W.; Wang, J.; and Han, J. 2014.
\newblock Entity linking with a knowledge base: Issues, techniques, and
  solutions.
\newblock \emph{IEEE Transactions on Knowledge and Data Engineering} 27(2):
  443--460.

\bibitem[{Wu et~al.(2019)Wu, Wang, Yuan, Wu, and Li}]{wu2019coreference}
Wu, W.; Wang, F.; Yuan, A.; Wu, F.; and Li, J. 2019.
\newblock Coreference resolution as query-based span prediction.
\newblock \emph{arXiv preprint arXiv:1911.01746} .

\bibitem[{Yamada et~al.(2020)Yamada, Washio, Shindo, and
  Matsumoto}]{yamada2020global}
Yamada, I.; Washio, K.; Shindo, H.; and Matsumoto, Y. 2020.
\newblock Global Entity Disambiguation with Pretrained Contextualized
  Embeddings of Words and Entities.
\newblock \emph{arXiv preprint arXiv:1909.00426} .

\bibitem[{Yang and Chang(2015)}]{yang2015s}
Yang, Y.; and Chang, M.-W. 2015.
\newblock S-MART: Novel Tree-based Structured Learning Algorithms Applied to
  Tweet Entity Linking.
\newblock In \emph{Proceedings of the 53rd Annual Meeting of the Association
  for Computational Linguistics and the 7th International Joint Conference on
  Natural Language Processing (Volume 1: Long Papers)}, 504--513.

\end{thebibliography}

\end{document}